\documentclass[letterpaper, 10 pt, conference]{ieeeconf}

\IEEEoverridecommandlockouts

\usepackage{amsmath,amsfonts}
\usepackage{amssymb}
\usepackage{bm}
\usepackage{algorithm}
\usepackage[noend]{algorithmic}
\usepackage{xcolor}
\usepackage{colortbl}
\usepackage{pifont}
\usepackage{array}
\usepackage{textcomp}
\usepackage{stfloats}
\usepackage{verbatim}
\usepackage{graphicx}
\usepackage{cite}
\usepackage[utf8]{inputenc}
\usepackage[T1]{fontenc}
\makeatletter
\let\NAT@parse\undefined
\makeatother
\usepackage{hyperref}
\hypersetup{colorlinks=false,citebordercolor={0 1 0},pdfborder={0 0 1}}
\usepackage{url}
\usepackage{booktabs}
\usepackage{nicefrac}
\usepackage{microtype}
\usepackage{subfigure}
\usepackage{multirow}
\usepackage{longtable}
\usepackage{threeparttable}
\usepackage{cuted}
\newcommand{\cmark}{\textcolor{green!60!black}{\ding{51}}}
\newcommand{\xmark}{\textcolor{red!70!black}{\ding{55}}}
\newcommand{\pmark}{\textcolor{orange!85!black}{\ding{115}}}
\newcommand{\makecell}[1]{\begin{tabular}{@{}c@{}}#1\end{tabular}}
\definecolor{locosusprow}{RGB}{230,240,248}

\title{\LARGE \bf
LocoSusp: Learning Virtual Suspension for Stable Wheeled-Legged Mobile Platforms
}

\author{Yongen Zhao$^{1,2}$, Xuanli Zhao$^{3}$, Jinyu Gao$^{4}$, Wenzhi Lu$^{1}$, Bo Peng$^{3}$, Pengwei Zhang$^{2,5}$,\\
Kailin Lyu$^{2,5}$, Hao Sun$^{6}$, Haifeng Wang$^{7}$, Zihao Xu$^{8}$, Kang Song$^{1}$, and Ce Hao$^{2}$%
\thanks{$^{1}$Yongen Zhao, Wenzhi Lu, and Kang Song are with Tianjin University, Tianjin, China.}%
\thanks{$^{2}$Yongen Zhao, Pengwei Zhang, Kailin Lyu, and Ce Hao are with Zhongguancun Academy, Beijing, China.}%
\thanks{$^{3}$Xuanli Zhao and Bo Peng are with DEEP Robotics, Hangzhou, China.}%
\thanks{$^{4}$Jinyu Gao is with Northwestern Polytechnical University, Xi'an, China.}%
\thanks{$^{5}$Pengwei Zhang and Kailin Lyu are with the Institute of Automation,
Chinese Academy of Sciences, Beijing, China.}%
\thanks{$^{6}$Hao Sun is with Guangzhou Automobile Group Co., Ltd., Guangzhou, China.}%
\thanks{$^{7}$Haifeng Wang is with Blovelight (Guangdong) Intelligent Technology Co., Ltd., Dongguan, China.}%
\thanks{$^{8}$Zihao Xu is with the National University of Singapore, Singapore.}%
}

\newcommand{\locosuspteaser}{%
\begingroup
\centering
\vspace{1.8em}
\refstepcounter{figure}
\label{fig:teaser}
\includegraphics[width=0.95\textwidth]{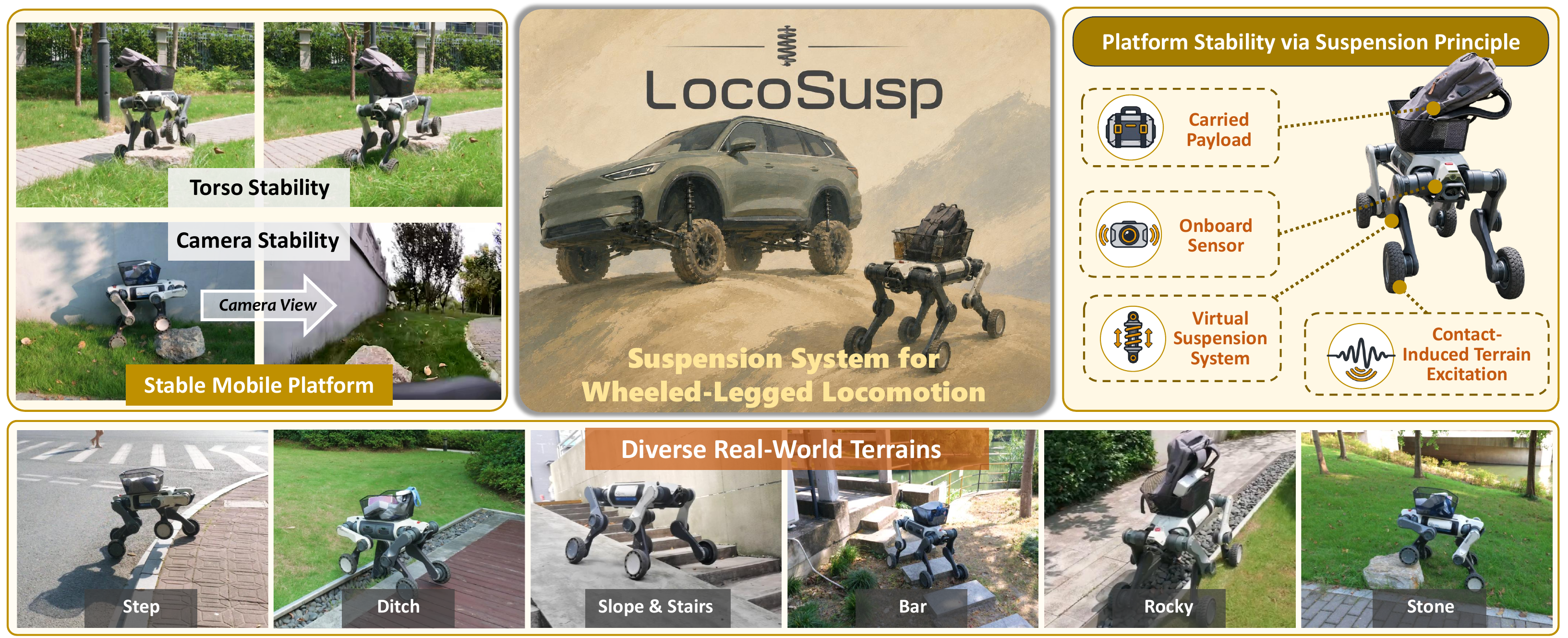}
\par\vspace{0.5em}

\noindent\parbox{0.96\textwidth}{\small \textbf{Fig.~\thefigure.} LocoSusp treats a wheeled-legged robot as a functional mobile suspension platform rather than only as a terrain traverser. The variable leg support absorbs terrain-induced vibration and impact while maintaining body, camera, and payload stability across diverse terrain.}
\par\vspace{1.0em}
\endgroup
}
\IEEEaftertitletext{\locosuspteaser}

\begin{document}

\maketitle
\thispagestyle{empty}
\pagestyle{empty}

\begin{abstract}
Wheeled-legged robots have achieved impressive terrain traversal capabilities, yet practical applications as mobile platforms require more than merely passing through challenging terrains. Terrain-induced vibration, impact, and body oscillation can significantly degrade onboard sensing and payload usability. We propose LocoSusp, a terrain-adaptive active suspension framework that reinterprets wheeled-legged robots as stable mobile platforms. LocoSusp exploits the suspension-like degrees of freedom provided by variable leg support and body configuration, and decomposes locomotion into a stability-preserving motion prior and terrain-dependent suspension adaptation. A contact-driven terrain excitation representation conditions residual compensation and support-height adjustment within a unified suspension policy. Experiments in simulation and on an M20 wheeled-legged robot demonstrate that LocoSusp improves platform-level stability across diverse terrains, achieving reduced vibration, improved body attitude stability, and enhanced camera/payload usability while maintaining locomotion performance. This work establishes a suspension-oriented perspective for wheeled-legged locomotion, moving beyond terrain traversal toward reliable mobile platform operation.
\end{abstract}

\section{Introduction}

Wheeled-legged robots have recently demonstrated remarkable mobility by combining the efficiency of rolling locomotion with the terrain adaptability of legged mechanisms. Existing studies have achieved impressive performance in challenging-terrain traversal, including outdoor navigation, perceptive locomotion, obstacle negotiation, and agile maneuvering \cite{lee2020challenging,miki2022perceptive,nahrendra2023dreamwaq,hoeller2024parkour,cheng2024extremeparkour,lee2024wheeledleggednav,zhao2026aware}. These studies mainly optimize robots as locomotion agents, where success is measured by traversability, command tracking, and recovery capability. In many real-world applications, wheeled-legged robots are expected to serve as mobile platforms carrying cameras, sensors, communication devices, or manipulation payloads. A stable mobile platform is required to maintain consistent body motion for onboard sensors and payloads.

For such applications, the objective extends beyond whether the robot can traverse the terrain to whether it can remain a stable and usable mobile platform while doing so. Yet existing wheeled-legged locomotion remains predominantly traversal-oriented, with limited consideration of vibration isolation, body stability, and sensor/payload usability.
Vehicle suspension control provides a mature reference for stabilizing mobile platforms under road disturbances \cite{tseng2015suspension}, but its methods cannot be directly transferred to wheeled-legged robots because of the substantial morphological and dynamical mismatch. Conventional vehicle suspension controllers are built around dedicated, relatively low-dimensional suspension mechanisms, whereas wheeled-legged robots rely on high-degree-of-freedom articulated legs simultaneously for locomotion, body support, and disturbance rejection, resulting in tightly coupled whole-body and contact dynamics. Nevertheless, these articulated degrees of freedom also allow the entire wheeled-legged body to be interpreted as a virtual suspension system. This calls for a suspension-oriented framework tailored to wheeled-legged morphology that can maintain platform stability and usability during traversal across diverse and complex terrains.

To address this problem, we propose \textbf{LocoSusp} (Locomotion with Suspension-oriented Control), a terrain-adaptive active suspension framework for stable wheeled-legged mobile bases. The key idea is to reinterpret wheeled-legged locomotion as a virtual suspension system, where terrain-induced excitation is sensed through robot-environment interaction and compensated through adaptive whole-body responses. LocoSusp learns a stability-preserving motion prior and a terrain-conditioned suspension adaptation module, using a contact-driven terrain excitation representation from proprioceptive histories to regulate vibration suppression and terrain-dependent ride-height adjustment.

We validate LocoSusp in IsaacLab simulation and on an M20 wheeled-legged robot across diverse terrains with different excitation characteristics, including rough surfaces, discrete impacts, slopes, and terrain transitions. The evaluation focuses on platform-level stability, including vertical vibration attenuation, body attitude stability, camera stability, and payload usability, while preserving locomotion performance. The results demonstrate that LocoSusp enables wheeled-legged robots to move beyond terrain traversal toward reliable mobile platform operation.

In summary, we identify our work with the following contributions:
\begin{itemize}
\item We propose LocoSusp, a terrain-adaptive active suspension framework that views wheeled-legged robots as stable mobile platforms beyond conventional terrain traversal.

\item We develop a hierarchical control framework that combines a stability-preserving motion backbone with terrain-conditioned adaptive compensation through contact-driven terrain excitation representation.

\item We validate LocoSusp through simulation and M20 real-robot experiments across diverse terrains, demonstrating improved body stability, and camera/payload usability while maintaining locomotion performance.
\end{itemize}

\section{Related Work}

\subsection{Legged and Wheeled-Legged Terrain Traversal}

Research on legged locomotion has addressed dynamic stability and terrain traversal through model predictive control, whole-body control, motion planning, and reinforcement learning. Model-based controllers regulate body motion, contact forces, foothold selection, and terrain-aware trajectories \cite{dicarlo2018convexmpc,bellicoso2017wbc,mastalli2020motionplanning,jenelten2022tamols}. Learning-based methods improve robustness and agility across natural terrain, slopes, stairs, gaps, sparse footholds, and external perturbations \cite{lee2020challenging,miki2022perceptive,jenelten2024dtc,hoeller2024parkour,cheng2024extremeparkour,hartmann2024deep,xiao2024paloco,xu2025facet,li2025sata,lyu2024rl2ac}. Wheeled-legged robots further combine rolling efficiency and legged adaptability. Representative methods optimize whole-body motion and online gait sequences under rolling contact constraints \cite{bjelonic2019keeprollin,bjelonic2020rollingdeep,bjelonic2021wbmpc}, and recent systems show robust outdoor navigation, blind stair climbing, energy-efficient agile rolling, and whole-body impedance coordination \cite{lee2024wheeledleggednav,chamorro2024blindstair,sun2025atros,shi2026wbcimpedance}. These methods improve mobility but prioritize traversal success, fall avoidance, command tracking, and dynamic feasibility. Successful traversal can still involve large vertical acceleration or camera shake, limiting sensing and payload usability. LocoSusp aims to improve platform stability by incorporating suspension-quality indicators into its control objectives.

\subsection{Terrain Representation and Adaptation}

Terrain-aware locomotion commonly uses exteroceptive terrain preview, proprioceptive history, or learned latent adaptation. Perceptive policies fuse depth or heightmap information with proprioception for robust locomotion in the wild \cite{miki2022perceptive,agarwal2023egocentric,he2025ame}. Proprioceptive adaptation methods such as RMA, DreamWaQ, and SLR infer latent environmental information from observation history and adapt policies to terrain, payload, or dynamics changes \cite{kumar2021rma,nahrendra2023dreamwaq,chen2025slr}. These representations are usually trained for traversability, command tracking, foothold feasibility, or robustness. LocoSusp focuses on terrain-induced body excitation to improve suspension quality and platform stability.

\subsection{Active Suspension and Platform Stability}

Vehicle active suspension systems estimate road disturbance and regulate ride height, damping, and body response to improve ride comfort, vibration isolation, and payload stability \cite{tseng2015suspension}. Mechanical design work for wheeled-legged rovers has also recognized active and passive compliance as important for rough-terrain interaction \cite{zhu2024stiffness}. However, wheeled-legged robot control has rarely treated the legs themselves as a virtual suspension layer for stable platform operation. LocoSusp transfers the suspension viewpoint into wheeled-legged locomotion: variable leg support is used to absorb terrain-induced excitation while preserving the nominal locomotion prior.

\section{Problem Formulation}

\begin{figure*}[!t]
\centering
\includegraphics[width=0.96\textwidth]{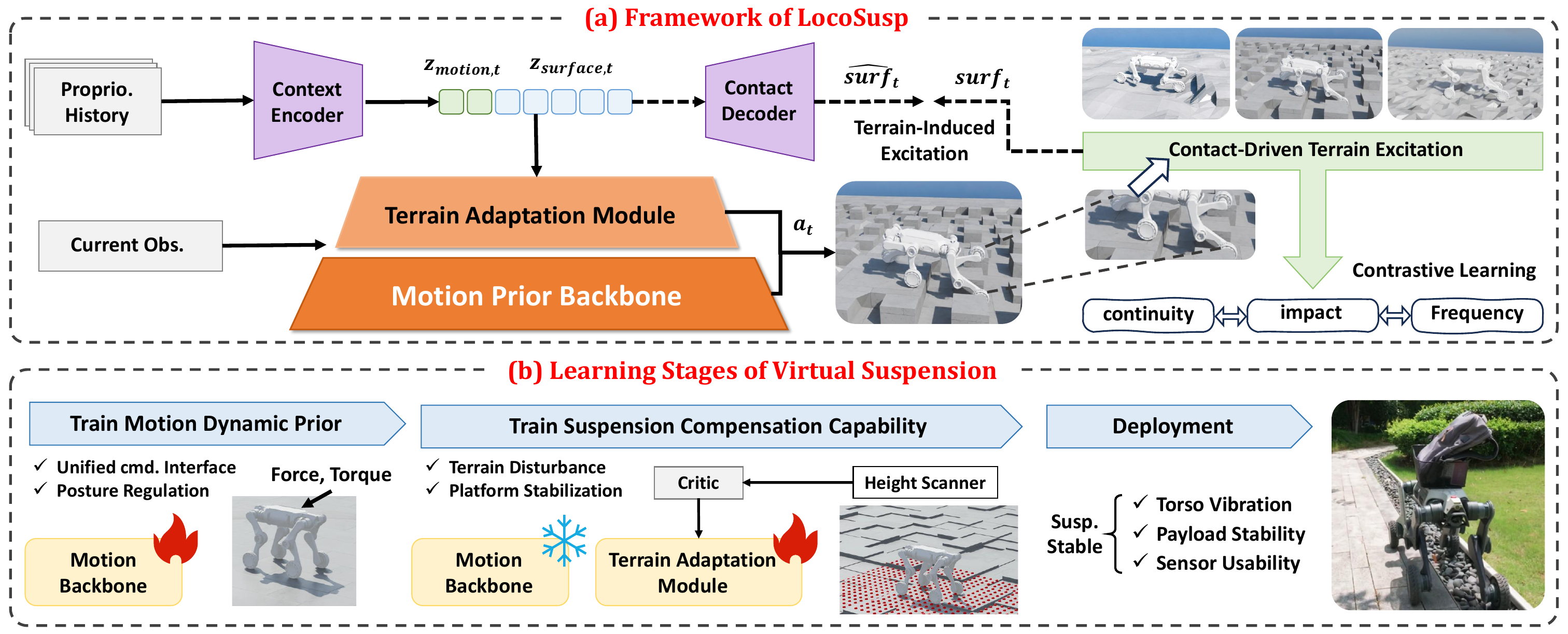}
\vspace{-0.5em}
\caption{Overview of LocoSusp. (a) LocoSusp formulates wheeled-legged locomotion as a suspension-oriented control problem, combining a stability-preserving motion prior with terrain-conditioned suspension adaptation driven by contact-induced excitation representation. (b) The hierarchical training procedure separates locomotion capability learning from suspension response learning, enabling stable mobile-platform operation across diverse terrains.}
\label{fig:locosusp_pipeline}
\end{figure*}

We consider a wheeled-legged robot commanded by a desired forward velocity and yaw rate $c_t=[v_t^{\mathrm{cmd}},\omega_t^{\mathrm{cmd}}]$. LocoSusp adopts a unified velocity-command interface, where different high-level motion objectives are converted into velocity references for the same low-level locomotion policy. Specifically, both constant-speed cruising and trajectory tracking employ a pure-pursuit-based velocity generation module~\cite{coulter1992implementation} to produce forward-velocity and yaw-rate commands, while teleoperation directly provides user-specified commands through the same interface.

\begingroup
\setlength{\intextsep}{6pt}
\setlength{\parskip}{0pt}
\begin{table}[H]
\centering
\makeatletter
\long\def\@makecaption#1#2{%
  {\centering\footnotesize #1\par\scshape #2\par}%
  \vskip 2pt\relax}
\makeatother
\caption{Functional comparison with related locomotion objectives. \cmark: supported, \pmark: partially supported, \xmark: not supported.}
\label{tab:related_function_comparison}
\begin{threeparttable}
\renewcommand{\arraystretch}{1.12}
\setlength{\tabcolsep}{1.8pt}
\scriptsize
\begin{tabular*}{\linewidth}{@{\extracolsep{\fill}}lccccc@{}}
\toprule
\textbf{Method family}
& \makecell{\textbf{Traversal}\\\textbf{success}}
& \makecell{\textbf{Command}\\\textbf{tracking}}
& \makecell{\textbf{Terrain}\\\textbf{latent}}
& \makecell{\textbf{Vibration}\\\textbf{objective}}
& \makecell{\textbf{Ride-height}\\\textbf{adaptation}} \\
\midrule
Model-based\tnote{a} & \cmark & \cmark & \pmark & \xmark & \pmark \\
Parkour RL\tnote{b} & \cmark & \pmark & \pmark & \xmark & \xmark \\
W-L WBC/MPC\tnote{c} & \cmark & \cmark & \pmark & \pmark & \pmark \\
Proprio. adapt.\tnote{d} & \cmark & \cmark & \cmark & \xmark & \xmark \\
\textbf{LocoSusp}\tnote{e} & \cmark & \cmark & \cmark & \cmark & \cmark \\
\bottomrule
\end{tabular*}
\vspace{0.5em}
\begin{tablenotes}[flushleft]
\scriptsize
\item[] \parbox{0.94\linewidth}{\textit{Notes.}
\textsuperscript{a} Model-based: convex MPC and motion planning~\cite{dicarlo2018convexmpc,mastalli2020motionplanning};
\textsuperscript{b} Parkour RL: traversal-oriented legged reinforcement learning~\cite{hoeller2024parkour,cheng2024extremeparkour};
\textsuperscript{c} W-L WBC/MPC: wheeled-legged whole-body and model predictive control~\cite{bjelonic2020rollingdeep,bjelonic2021wbmpc};
\textsuperscript{d} Proprio. adapt.: proprioceptive terrain/dynamics adaptation~\cite{kumar2021rma,nahrendra2023dreamwaq};
\textsuperscript{e} LocoSusp: our suspension-quality-aware wheeled-legged framework.}
\end{tablenotes}
\end{threeparttable}
\end{table}

Unlike conventional locomotion objectives that only consider whether the robot reaches the target, we formulate wheeled-legged locomotion as a coupled mobility--stability optimization problem. Given command $c_t=[v_t,\omega_t]$, the policy should maximize traversal capability while minimizing terrain-induced platform excitation, $\pi^*=\arg\min_{\pi}\mathbb{E}\sum_t\gamma^t\left(J_{\mathrm{track}}+\lambda_sJ_{\mathrm{susp}}\right)$, where $J_{\mathrm{susp}}=w_a a_z^2+w_R\lVert e_{rp}\rVert_2^2+w_j\dot{a}_z^2$ measures suspension quality.
This formulation makes the work's core distinction explicit: \emph{traversability} measures whether the robot passes the terrain, while \emph{suspension quality} measures whether the body remains a usable mobile platform during traversal.
\par\endgroup

During steady rolling on smooth, level terrain, the desired nominal behavior is to maintain a stable body--leg configuration at a prescribed support height with sustained four-wheel contact. Under terrain-induced excitation, the control objective is to attenuate body vibration and attitude disturbances through adaptive leg motion while maintaining command tracking and feasible support. The support-height reference may also be adjusted across terrain conditions to complement disturbance compensation.

\section{Method}

\subsection{Overview}

Unlike conventional vehicles where suspension elements are mechanically separated from the chassis, wheeled-legged robots possess redundant leg-body DoFs that can be exploited as a virtual suspension layer.

LocoSusp combines a learned stability-preserving motion prior with terrain-conditioned residual compensation. The motion prior provides nominal locomotion and support-height regulation, while the residual policy learns corrective actions conditioned on contact-driven terrain excitation:
\begin{align}
a_t^{\mathrm{base}}
&=\pi_{\mathrm{base}}(o_t,c_t,h_t^*),\\
\Delta a_t^{\mathrm{terrain}}
&=\pi_{\mathrm{res}}(o_{t-H:t},z_t^{\mathrm{terr}}),\\
a_t&=a_t^{\mathrm{base}}+\Delta a_t^{\mathrm{terrain}}.
\label{eq:locosusp_action}
\end{align}
Here $h_t^*$ is the desired ride height, $z_t^{\mathrm{terr}}$ is the contact-driven terrain excitation latent, and $H$ is the observation-history length. The design intentionally preserves a stable body configuration on flat and smooth terrain while allowing stronger suspension adaptation under rough, discontinuous, or high-impact contact.

\subsection{Contact-Driven Terrain Excitation}

\begin{figure}[t]
\centering
\includegraphics[width=\linewidth]{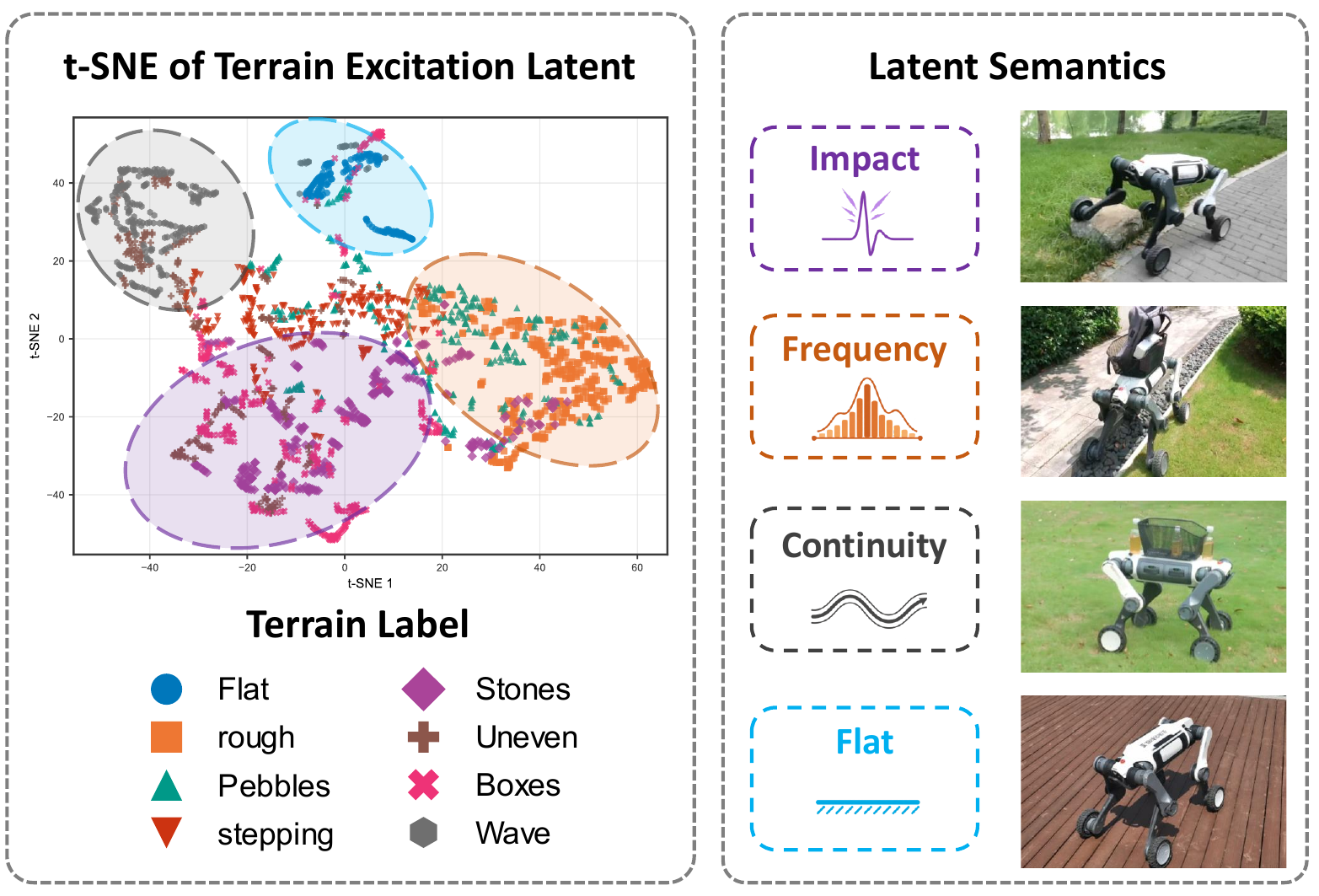}
\vspace{-1.1em}
\caption{Visualization of the learned terrain-representation latent over different terrain types. The separated latent structure supports the design goal that contact-driven proprioceptive histories encode terrain excitation characteristics relevant to suspension adaptation, rather than only memorizing commanded motion.}
\label{fig:terrain_latent_visualization}
\vspace{-1.2em}
\end{figure}

Terrain excitation is difficult to observe directly. Dense geometric reconstruction may be expensive, and visual terrain appearance does not always predict the disturbance transmitted through wheel-ground contact. LocoSusp therefore estimates terrain excitation from proprioceptive histories. Let $q^w_{z,t-H:t}$ denote the vertical wheel/foot position history over the last $H$ frames. We compute a normalized spectrum:

\begingroup
\small
\begin{equation}
P_{i,t,k} = \left| \mathrm{FFT}_{\mathrm{time}} \left( q_{z,i,t-H:t}^{w} \right)_{k} \right|^{2},
\end{equation}
\endgroup
\begingroup
\small
\begin{equation}
\bar{P}_{i,t,k} = \frac{P_{i,t,k}}{\displaystyle \sum_{j=0}^{K-1} P_{i,t,j} + \epsilon},
\end{equation}
\endgroup
For each wheel $i$, we independently apply the FFT along the temporal axis to its vertical position history and normalize the resulting power spectrum. The four features in Table~\ref{tab:fft_features} are extracted separately for each wheel and concatenated in a fixed wheel order, yielding a 16-dimensional terrain-excitation target $\mathbf{F}_t$,
\begin{equation}
\mathbf{f}_{i,t}
=
\left[
f_{\mathrm{peak},i,t},
e_{\mathrm{peak},i,t},
h_{f,i,t},
E_{exc,i,t}
\right]^{\top}
\in\mathbb{R}^{4},
\label{eq:wheel_fft_features}
\end{equation}
\begin{equation}
\mathbf{F}_{t}
=
\mathrm{Concat}\left(\mathbf{f}_{1,t},\mathbf{f}_{2,t},\mathbf{f}_{3,t},\mathbf{f}_{4,t}\right)
=
\begin{bmatrix}
\mathbf{f}_{1,t}\\
\mathbf{f}_{2,t}\\
\mathbf{f}_{3,t}\\
\mathbf{f}_{4,t}
\end{bmatrix}
\in\mathbb{R}^{16}.
\label{eq:terrain_excitation_target}
\end{equation}
Rather than treating terrain types as mutually exclusive categories, we organize robot--terrain interactions along three complementary excitation characteristics: impact intensity, frequency content, and temporal continuity (Fig.~\ref{fig:terrain_latent_visualization}). This excitation-oriented view guides the construction of the terrain suite and the interpretation of contact-driven observations.
To encourage a structured terrain-excitation latent space, we further introduce a label-guided contrastive regularization. Samples with the same terrain excitation category are treated as positive pairs, while different categories are pushed apart.

\begingroup
\small
\begin{equation}
\mathcal{L}_{\mathrm{con}}
= -\sum_i \frac{1}{|P(i)|}\sum_{j\in P(i)}
\log
\frac{\exp(z_i^{\mathrm{terr}}\cdot z_j^{\mathrm{terr}}/\tau)}
{\sum_{k\neq i}\exp(z_i^{\mathrm{terr}}\cdot z_k^{\mathrm{terr}}/\tau)}.
\label{eq:contrastive_loss}
\end{equation}
\endgroup
where, $P(i)=\{j\mid y_j=y_i,\ j\neq i\}$ denotes the positive set for sample $i$, containing other samples with the same terrain-excitation label $y_i$.

\begin{table}[t]
\caption{Contact-driven FFT terrain-excitation features.}
\label{tab:fft_features}
\centering
\scriptsize
\setlength{\tabcolsep}{3pt}
\begin{tabular}{@{}p{0.25\columnwidth}>{\centering\arraybackslash}p{0.53\columnwidth}>{\centering\arraybackslash}p{0.16\columnwidth}@{}}
\toprule
Feature & \multicolumn{1}{c}{Definition} & Symbol \\
\midrule
Peak frequency & $\displaystyle \frac{1}{K-1}\arg\max_k P_k$ & $f_{\mathrm{peak}}$ \\
Peak energy & $\displaystyle \max_k \bar{P}_k$ & $e_{\mathrm{peak}}$ \\
Spectral entropy & $\displaystyle -\frac{1}{\log K}\sum_k \bar{P}_k\log(\bar{P}_k+\epsilon)$ & $h_f$ \\
Excitation energy & $\displaystyle \log\!\left(\frac{1}{K}\sum_{k=0}^{K-1}P_{k}+\epsilon\right)$ & $E_{exc}$ \\
\bottomrule
\end{tabular}
\end{table}

A history encoder $E_{\theta}$ receives the last $H=10$ frames of proprioceptive observation and command history, {\scriptsize $[z_t^{\mathrm{terr}},z_t^{\mathrm{mot}}]=E_{\theta}(o_{t-H:t},c_{t-H:t})$}.
Two auxiliary decoders reconstruct terrain-excitation and motion features, $\hat{\mathbf{F}}_t=D_{\mathrm{terr}}(z_t^{\mathrm{terr}})$ and $\hat{u}_t=D_{\mathrm{mot}}(z_t^{\mathrm{mot}})$.
Here $u_t=[v_{x,t},\dot{\psi}_t]$. The representation loss is

\begingroup
\small
\begin{equation}
\mathcal{L}_{\mathrm{repr}}
= \lambda_{\mathrm{terr}}\|\hat{F}_t-F_t\|_2^2
+ \lambda_{\mathrm{mot}}\|\hat{u}_t-u_t\|_2^2
+ \lambda_c\mathcal{L}_{\mathrm{con}}.
\label{eq:representation_loss}
\end{equation}
\endgroup

\subsection{Stability Backbone}

The stability backbone learns command tracking, support-height regulation, and disturbance rejection on flat and low-complexity terrain. Stand-up and lie-down recordings provide nominal leg-joint configurations at three reference heights. To encourage height-consistent support configurations, we add a height-conditioned Mahalanobis posture penalty to the locomotion reward:

\begingroup
\small
\begin{equation}
\begin{gathered}
r_t^{\mathrm{post}}
= -w_q\,\mathbf{e}_{q,t}^{\top}
\left(\Sigma_q(h_t^\ast)+\epsilon I\right)^{-1}
\mathbf{e}_{q,t},\\
\mathbf{e}_{q,t}
= \mathbf{q}_t-\mu_q(h_t^\ast).
\end{gathered}
\label{eq:posture_reward}
\end{equation}
\endgroup
Here, $\mathbf{q}_t$ contains leg-joint positions, $\mu_q$ and $\Sigma_q$ denote the reference configuration and covariance associated with the commanded height, and $\epsilon$ regularizes the inverse. This soft penalty discourages deviations from the nominal configuration while allowing corrective motion. Randomized external body forces and torques are applied during training to promote robust command tracking and support-height regulation~\cite{campanaro2022minimalrandomization}, providing a reliable motion prior for subsequent terrain-conditioned suspension adaptation.

\subsection{Adaptive Residual Suspension}

\begin{figure}[t]
\centering
\includegraphics[width=0.88\linewidth]{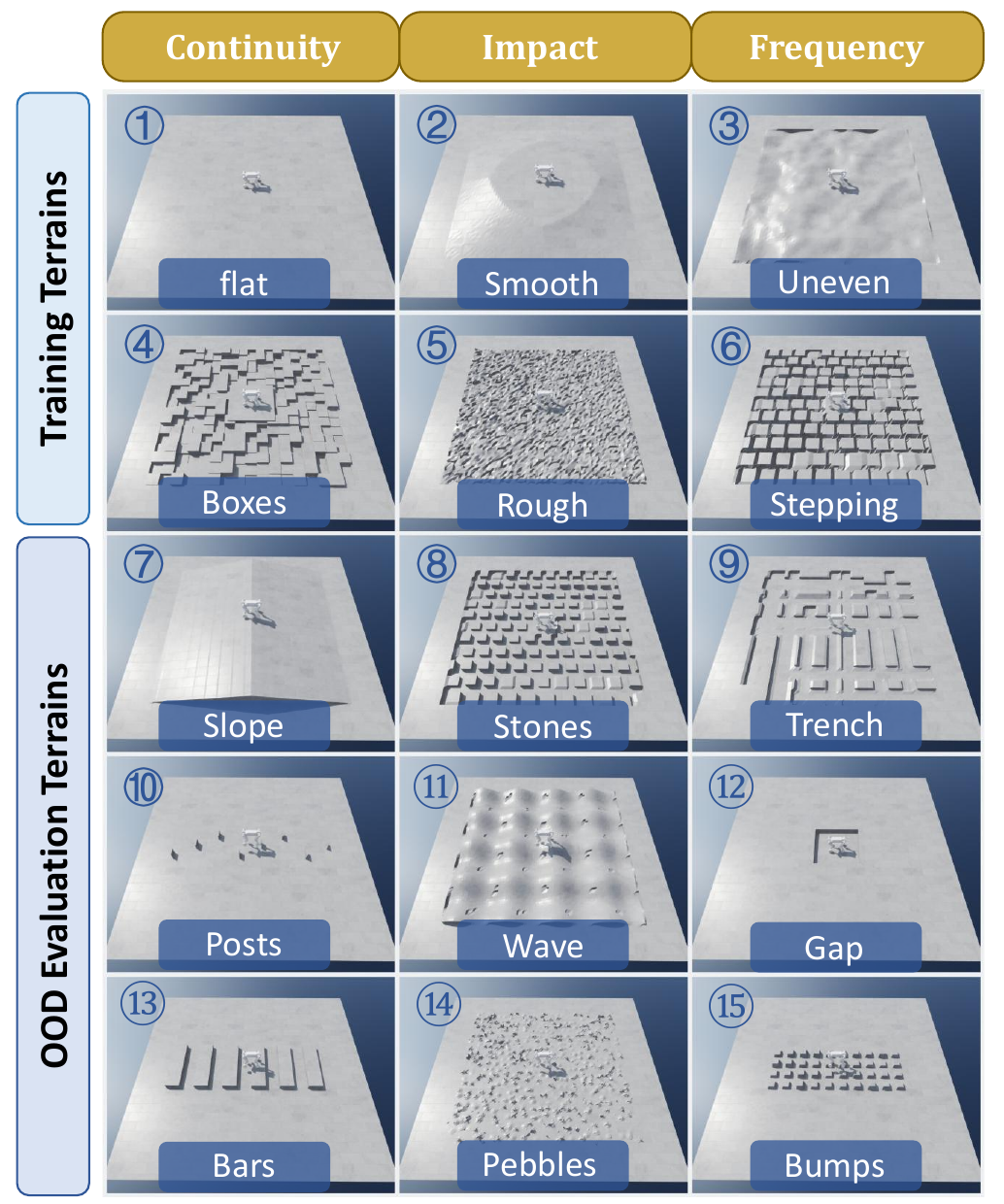}
\vspace{-0.4em}
\caption{Terrain classification used for training and validation. The terrain set is organized by three complementary excitation characteristics and platform-stability challenge.}
\label{fig:terrain_classification}
\vspace{-1.2em}
\end{figure}

After training the backbone, we freeze it and train the residual suspension policy over terrain classes with different excitation characteristics, including flat, smooth, uneven, boxes, rough, and stepping (Fig.~\ref{fig:terrain_classification}). The residual policy receives the terrain latent and outputs a compensation action $\Delta a_t^{\mathrm{terrain}}=\pi_{\mathrm{res}}(o_{t-H:t},z_t^{\mathrm{terr}})$.
To avoid residual interference on smooth terrain, we store latent prototypes for flat and smooth-slope terrain, $\mathcal{P}_{\mathrm{sm}}=\{z^{\mathrm{flat}},z^{\mathrm{smooth}}\}$. The smooth-terrain similarity is
\begin{equation}
s_t^{\mathrm{sm}}=\max_{z\in\mathcal{P}_{\mathrm{sm}}}
\frac{z_t^{\mathrm{terr}}\cdot z}{\|z_t^{\mathrm{terr}}\|\|z\|+\epsilon}.
\end{equation}
The residual penalty increases when the latent is close to smooth terrain, with $\lambda_t=\lambda_{\max}\sigma(\beta(s_t^{\mathrm{sm}}-\delta_{\mathrm{sm}}))$, and the regularization is $\mathcal{L}_{\mathrm{res-reg}}=\lambda_t\|\Delta a_t^{\mathrm{terrain}}\|_2^2$.
This barrier-like regularization keeps the residual nearly inactive on flat/smooth terrain and allows stronger suspension compensation on rough or impact-rich terrain.

\textbf{Platform stabilization objective.}
The residual policy is trained to reduce vertical body acceleration and roll--pitch deviations while retaining command tracking and locomotion feasibility. We define the stabilization reward as:

\begingroup
\small
\begin{equation}
r_t^{\mathrm{stab}}
= -w_a\left(\frac{a_{z,t}^{\mathrm{lin}}}{s_a}\right)^2
-w_R\left\|\frac{\bm{e}_{rp,t}}{s_R}\right\|_2^2,
\label{eq:stab_reward}
\end{equation}
\begin{equation}
\bm{e}_{rp,t}
=
\begin{bmatrix}
\phi_t-\phi_t^{\mathrm{ref}}\\
\theta_t-\theta_t^{\mathrm{ref}}
\end{bmatrix}.
\label{eq:rp_error}
\end{equation}
\endgroup
Here $a_{z,t}^{\mathrm{lin}}$ is the world-vertical linear acceleration with gravity removed, $\bm{e}_{rp,t}$ is the roll/pitch error relative to the target posture, and $s_a$ and $s_R$ normalize the acceleration and attitude scales. These terms are combined with the locomotion objectives and smooth-terrain residual regularization. The full reward specification and weights are provided in the supplementary material.

\subsection{Support-Height Adaptation}

Support-height adaptation is learned jointly with residual suspension control. An additional scalar output of the residual policy specifies a target support height within $[0.50,0.55]$ m. The target is rate-limited and passed to the frozen, height-conditioned backbone, while the remaining outputs provide direct action compensation.

\section{Experiments}

\begin{figure}[t]
\centering
\includegraphics[width=\linewidth]{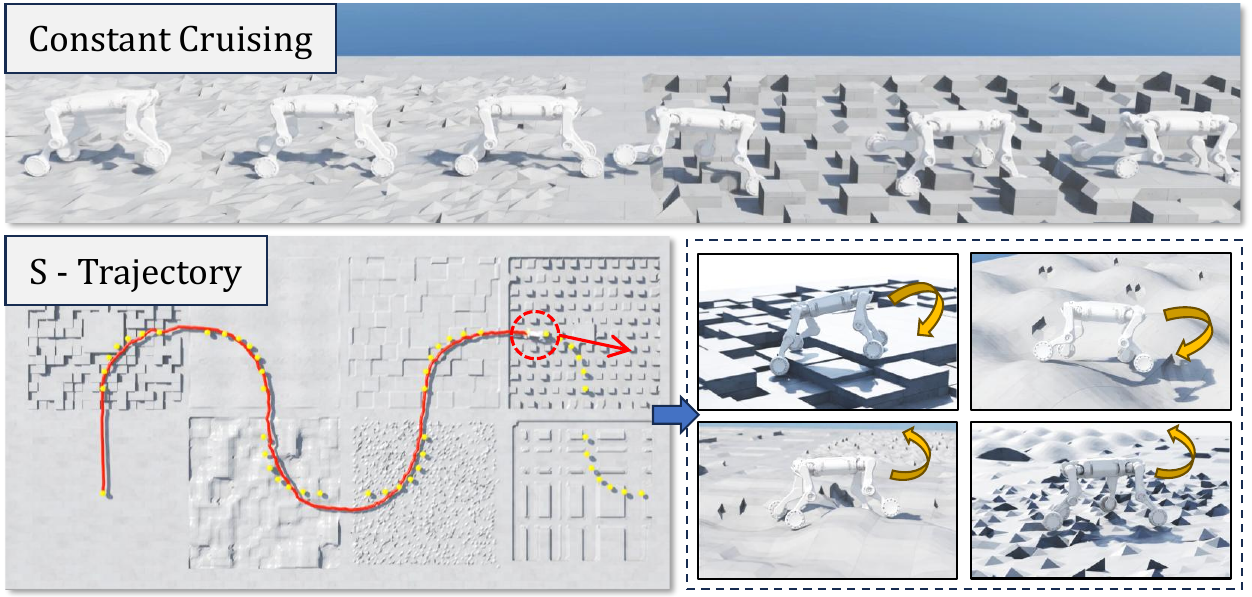}\\[0.6em]
{\scriptsize
\setlength{\tabcolsep}{4pt}
\renewcommand{\arraystretch}{0.95}
\begin{tabular*}{\linewidth}{@{\extracolsep{\fill}}lccc@{}}
\toprule
Task & Commands & Terrain IDs & Speed (m/s) \\
\midrule
Constant cruising & $v_x,\ \omega_z$ & 1--6, 7--10, 13, 15 & 1.0--2.5 \\
S-trajectory & $v_x,\ \omega_z$ & 1--6, 9, 11, 12, 14 & 0.5--1.5 \\
\bottomrule
\end{tabular*}}
\vspace{-0.3em}
\caption{Simulation routes used for evaluation. Constant-speed cruising traverses multiple terrain types along a straight route, while S-trajectory tracking evaluates curved-path command tracking over representative terrains.}
\label{fig:simulation_routes}
\vspace{-1.2em}
\end{figure}

We evaluate LocoSusp in IsaacLab simulation \cite{mittal2025isaaclab} and on an M20 wheeled-legged robot across terrains with diverse excitation characteristics. Simulation experiments include constant-speed cruising and S-trajectory tracking (Fig.~\ref{fig:simulation_routes}), while real-robot demonstrations cover representative outdoor terrains and qualitative teleoperation under irregular commands.

Our evaluation addresses three questions: (Q1) Can LocoSusp improve platform stability across diverse terrains while maintaining traversal success and command-tracking performance? (Q2) Does contact-driven terrain excitation provide an effective representation for triggering terrain-dependent suspension adaptation? (Q3) Does the learned suspension behavior remain effective on real hardware in supporting body, camera, and payload stability?

\subsection{Experimental Setup and Evaluation Protocol}

We compare with DreamWaQ \cite{nahrendra2023dreamwaq} and PA-LOCO \cite{xiao2024paloco}. We also include ablations: \emph{w/o residual}, \emph{w/o terrain excitation}, \emph{average terrain excitation}, \emph{w/o residual regularization}, and \emph{w/o support-height adaptation}. The \emph{w/o terrain excitation} variant removes $z_t^{\mathrm{terr}}$ from the residual-policy and ride-height conditioning inputs. In contrast, \emph{average terrain excitation} replaces the online latent with the fixed global mean $\bar{z}^{\mathrm{terr}}$ over the training set, preserving an average suspension response while disabling online excitation adaptation.

\begin{table*}[t]
\centering
\caption{Quantitative comparison in simulation for constant-speed cruising and S-trajectory tracking.}
\label{tab:sim_results}
\resizebox{\textwidth}{!}{
\begin{tabular}{l|ccccc|ccccc}
\toprule
& \multicolumn{5}{c|}{Constant-Speed Cruising}
& \multicolumn{5}{c}{S-Trajectory Tracking} \\
\cmidrule(lr){2-6}\cmidrule(lr){7-11}
Method
& SR$\uparrow$ & RMS $a_z\downarrow$ & RMS p/r$\downarrow$ & $T_{\mathrm{vib}}\downarrow$ & $J_{\mathrm{cam}}\downarrow$
& SR$\uparrow$ & Track err.$\downarrow$ & RMS $a_z\downarrow$ & RMS p/r$\downarrow$ & $T_{\mathrm{vib}}\downarrow$ \\
\midrule
DreamWaQ~\cite{nahrendra2023dreamwaq} & 90.1 & 4.43 & 5.52 & 0.15 & 16.1 & 91.3 & 0.67 & 2.78 & 4.59 & 0.21 \\
PA-LOCO~\cite{xiao2024paloco} & 60.8 & 8.38 & 8.18 & 0.24 & 34.3 & 87.1 & 1.11 & 2.91 & 5.92 & 0.29 \\
w/o residual & 85.7 & 6.21 & 7.86 & 0.21 & 29.8 & 82.3 & 0.65 & 3.45 & 6.13 & 0.44 \\
w/o terrain excitation & 89.4 & 5.12 & 6.78 & 0.18 & 24.6 & 88.5 & 0.58 & 2.81 & 5.24 & 0.32 \\
Avg. terrain excitation & 91.0 & 4.83 & 6.42 & 0.15 & 22.1 & 91.7 & 0.55 & 2.62 & 4.98 & 0.28 \\
w/o height adapt. & 92.1 & 4.04 & 6.01 & 0.11 & 15.2 & 92.9 & 0.53 & 2.30 & 4.57 & 0.23 \\
\rowcolor{locosusprow}
\textbf{LocoSusp} & \textbf{94.1} & \textbf{3.76} & \textbf{5.22} & \textbf{0.095} & \textbf{14.4} & \textbf{95.6} & \textbf{0.506} & \textbf{2.11} & \textbf{4.09} & \textbf{0.19} \\
\bottomrule
\end{tabular}}
\vspace{-0.8em}

\end{table*}

We report both mobility and suspension-quality metrics. Mobility metrics include success rate (SR), tracking error, and traversal speed. Platform metrics include RMS vertical acceleration, RMS pitch/roll, impact peak, vertical vibration transmissibility, and camera/payload stability. For a trajectory of length $T$, we compute $\mathrm{RMS}(a_z)=\sqrt{\frac{1}{T}\sum_{t=1}^{T}(a_{z,t}-\bar{a}_z)^2}$ and $\mathrm{RMS}(\theta,\phi)=\sqrt{\frac{1}{T}\sum_{t=1}^{T}(\theta_t^2+\phi_t^2)}$. For constant-speed cruising, we additionally report a camera-stability jitter metric\par
\begingroup
\small
\begin{equation}
J_{\mathrm{cam}}=
\sqrt{
\frac{1}{T}\sum_{t=1}^{T}
\left[
\operatorname*{median}_{i\in\mathcal{I}_t}
\left\|p_i^t-\Pi\!\left(H_t p_i^{t-1}\right)\right\|_2
\right]^2
},
\end{equation}
\endgroup
where $p_i^t$ is the image position of tracked feature $i$, $\mathcal{I}_t$ is the valid matched-feature set, $H_t$ is the frame-to-frame homography, and $\Pi(\cdot)$ denotes image-plane projection. Lower $J_{\mathrm{cam}}$ indicates smaller homography-compensated camera jitter.
Vertical vibration power ratio is computed as
\begin{equation}
T_{\mathrm{vib}}=
\frac{
\sum_{f\in\Omega}\left|\operatorname{FFT}(a_z^{\mathrm{base}})(f)\right|^2
}{
\frac{1}{4}\sum_{i=1}^{4}
\sum_{f\in\Omega}\left|\operatorname{FFT}(a_{wz,i})(f)\right|^2+\epsilon
},
\end{equation}
where $\Omega$ is the target vibration band and $a_{wz,i}$ is the vertical acceleration of wheel $i$. Lower transmissibility means better suspension isolation.

\subsection{Simulation Results}

\begin{figure}[t]
\centering
\includegraphics[width=\linewidth]{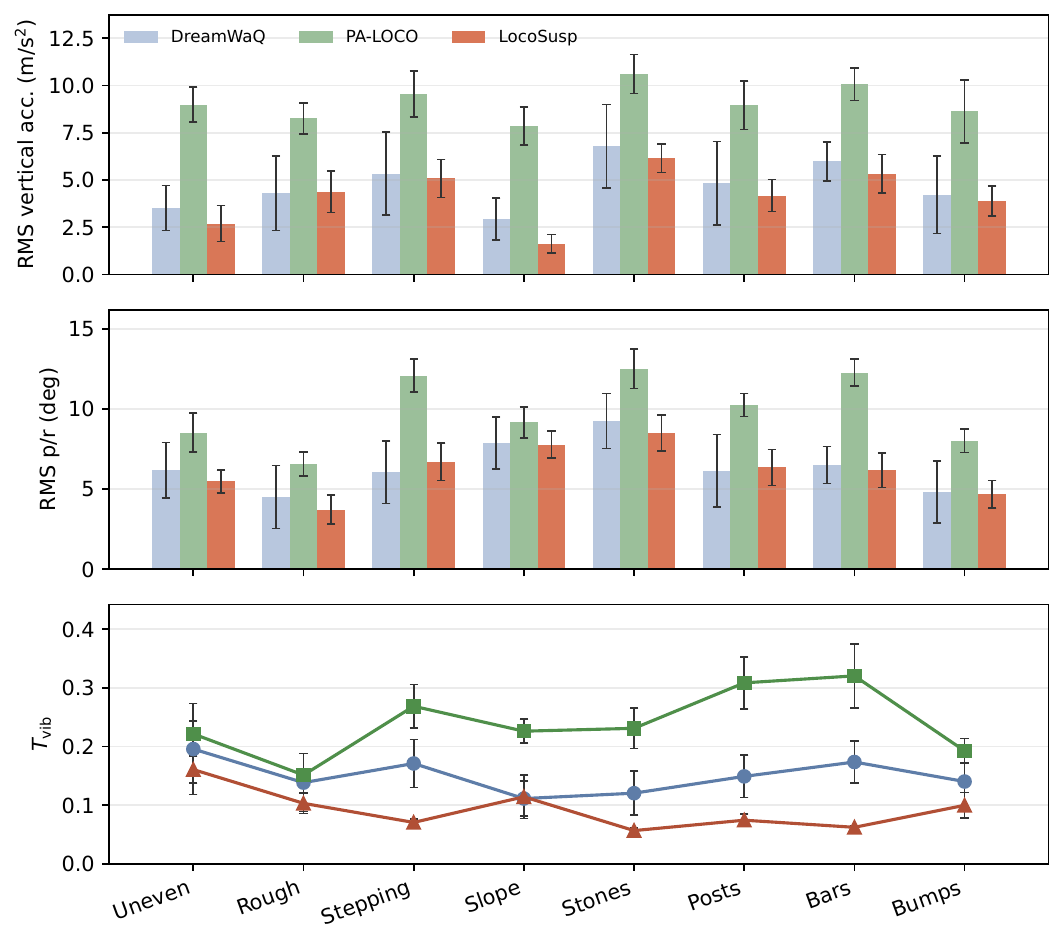}
\vspace{-1.2em}
\caption{Per-terrain simulation comparison under constant-speed cruising. The three panels report RMS vertical acceleration, RMS pitch/roll, and vertical vibration transmissibility over representative terrain labels.}
\label{fig:baseline_main_results}
\vspace{-1.8em}
\end{figure}

\begin{figure}[htbp]
\centering
\includegraphics[width=\linewidth]{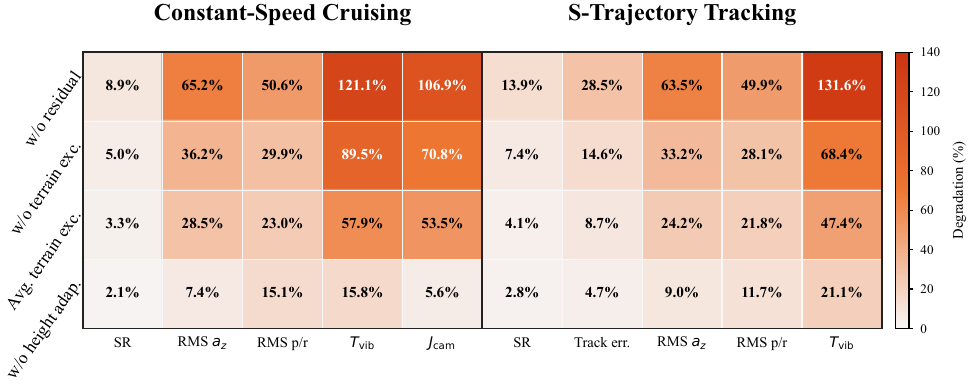}\\[0.3em]
\includegraphics[width=0.88\linewidth]{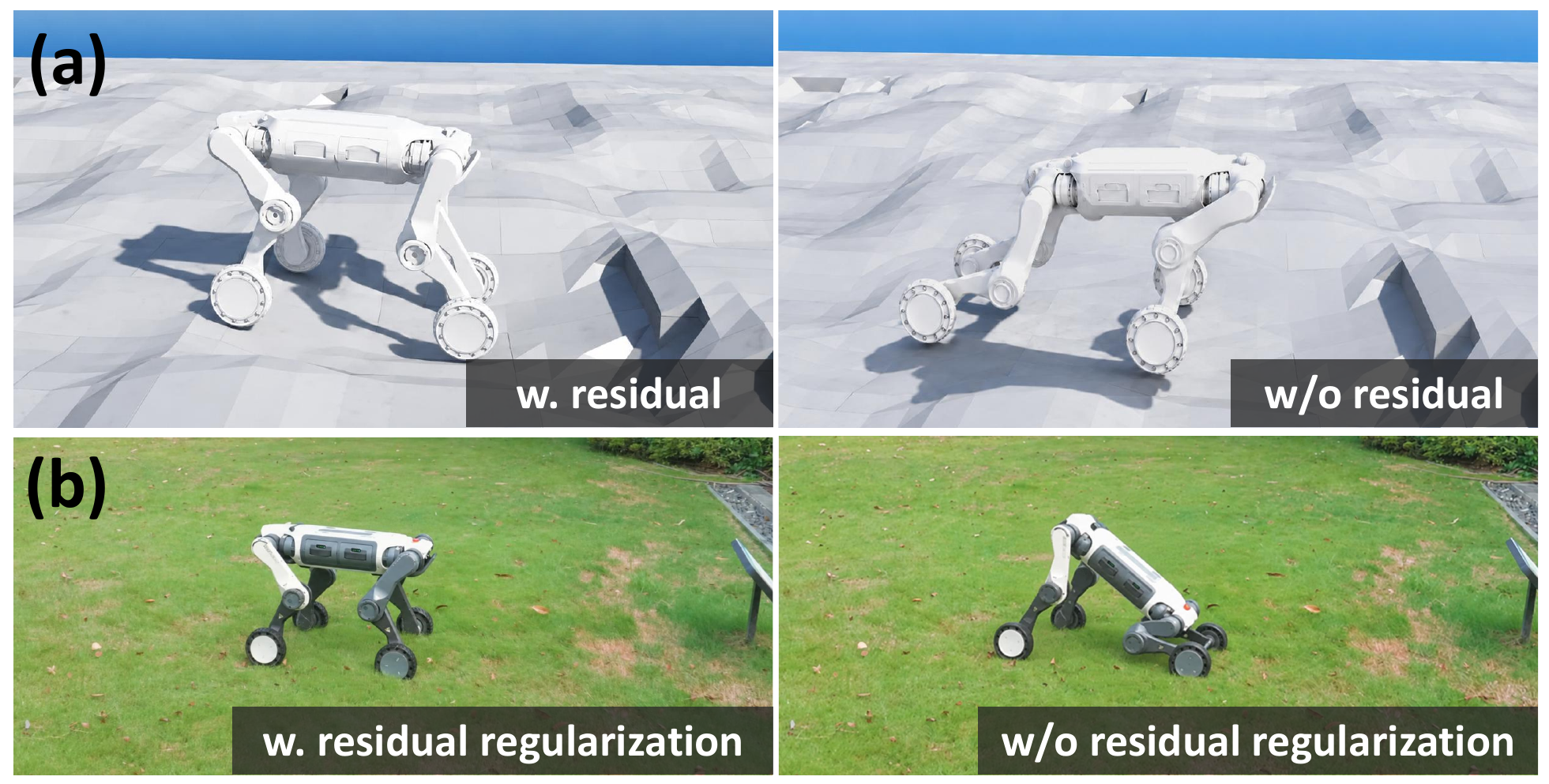}
\vspace{-0.4em}
\caption{Ablation summary showing the contribution of each LocoSusp component. (a) Body–leg configuration comparison between LocoSusp and the single-policy variant (w/o residual) in multi-terrain training. (b) Without residual regularization, the robot tends to tremble on smooth terrain or while standing still until it falls.}
\label{fig:ablation_summary}
\vspace{-1.8em}
\end{figure}

Table~\ref{tab:sim_results} reports the main simulation comparison under different locomotion commands. For constant-speed cruising, LocoSusp achieves higher traversal success while reducing vertical acceleration, pitch/roll disturbance, vibration transmissibility, and camera jitter compared with both baselines. For S-trajectory tracking, LocoSusp maintains comparable command tracking performance while achieving lower platform disturbance, indicating that suspension-oriented stabilization does not compromise locomotion capability.

Figure~\ref{fig:baseline_main_results} further analyzes the stability response across individual terrain conditions using three suspension-quality metrics: RMS vertical acceleration, RMS pitch/roll, and vertical vibration transmissibility. LocoSusp consistently reduces terrain-induced body excitation across uneven, slope, stone-like, and bump-like terrains, demonstrating that the learned virtual suspension response adapts to different disturbance characteristics rather than specific terrain types.

\begin{figure}[t]
\centering
\includegraphics[width=\linewidth]{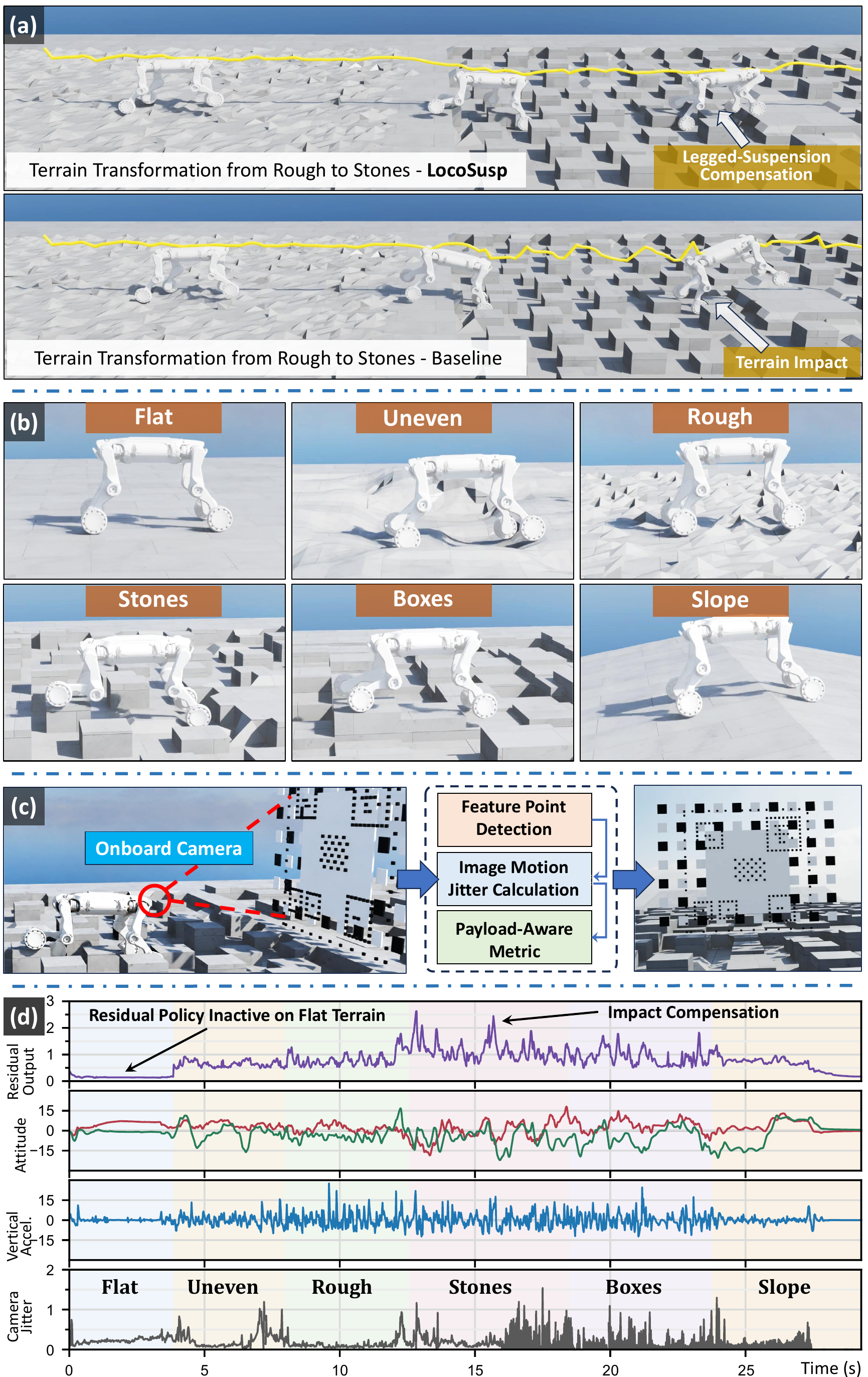}
\vspace{-1.5em}
\caption{Multi-terrain cruising evaluation for suspension-oriented locomotion. The robot traverses a 1$\times$N terrain sequence containing diverse excitation patterns, including smooth surfaces, continuous roughness, discrete impacts, and terrain transitions. (a) Overall cruising route and terrain transition sequence used for evaluation. (b) Representative terrain segments covering different disturbance characteristics. (c) Illustration of the camera jitter metric, which quantifies platform stability from onboard visual motion rather than only body-state measurements. (d) Representative temporal responses showing residual suspension adaptation, body vibration, and camera stability during traversal.}
\label{fig:simulation_env}
\vspace{-1.8em}
\end{figure}

Figure~\ref{fig:simulation_env} further shows representative simulation scenes used to evaluate LocoSusp under diverse terrain-induced excitation patterns. The evaluation consists of a 1$\times$N multi-terrain cruising sequence with continuous terrain transitions, representative terrain segments, and onboard-camera-based stability analysis. This unified evaluation enables us to examine whether the terrain excitation representation and residual suspension policy can adaptively regulate body motion and maintain platform stability beyond nominal traversability. Compared with the baseline, LocoSusp achieves smoother body motion throughout the multi-terrain sequence, especially during terrain transitions where disturbance characteristics change rapidly. 

\subsection{Ablation Study}


The ablations investigate the contribution of each component to the proposed suspension mechanism. Figure~\ref{fig:ablation_summary} summarizes how different variants affect stability-oriented metrics over all simulated terrains. The \emph{w/o residual} variant exhibits larger body vibration and attitude disturbances, demonstrating that decomposing locomotion and suspension adaptation enables more effective platform stabilization.
The \emph{w/o terrain excitation} variant removes the explicit terrain excitation representation, while \emph{average terrain excitation} preserves a generic excitation prior but lacks online adaptation to changing terrain disturbances. Their comparison with the full online latent verifies that suspension responses should be conditioned on instantaneous contact-induced excitation.
The \emph{w/o residual regularization} variant causes unnecessary residual actions on flat terrain and leads to over-active suspension responses when the nominal policy is already stable. Finally, the \emph{w/o support-height adaptation} variant increases impact responses during terrain transitions, indicating that posture regulation provides complementary stabilization effects.


\subsection{Real-World Validation of Suspension Behavior}

\begin{figure}[htbp]
\centering
\vspace{-0.6em}
\includegraphics[width=0.95\linewidth]{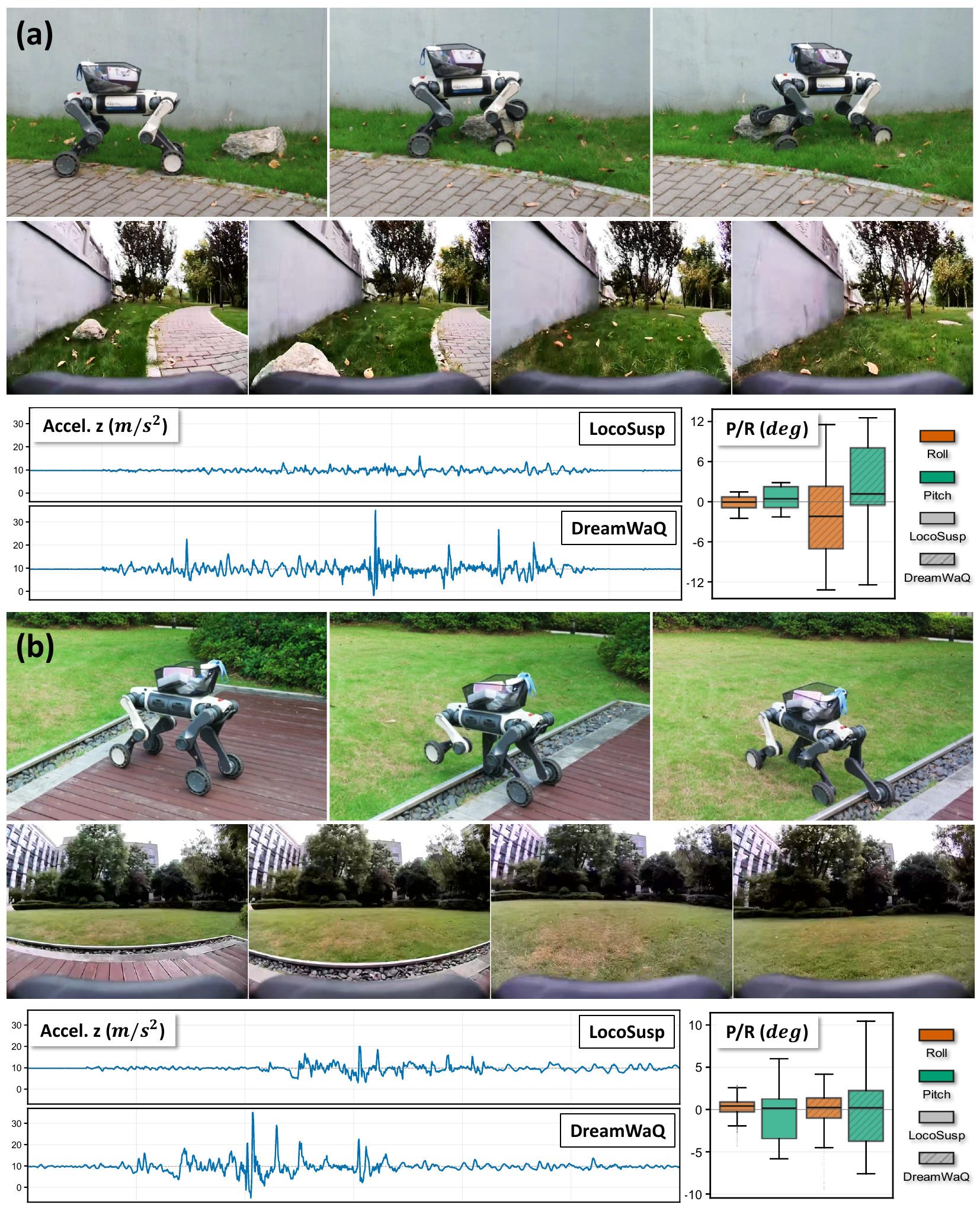}
\vspace{-0.5em}
\caption{Real-world validation of excitation-aware suspension behavior on the M20 wheeled-legged platform. (a) Traversal of an isolated rock obstacle producing discrete impact excitation. (b) Transition from a raised path to grass producing continuous terrain transition excitation. The onboard camera views and body motion measurements demonstrate the effectiveness of virtual suspension adaptation under real contact uncertainty.}
\label{fig:real_robot_with_data}
\vspace{-1.0em}
\end{figure}

\begin{figure}[htbp]
\centering
\includegraphics[width=\linewidth]{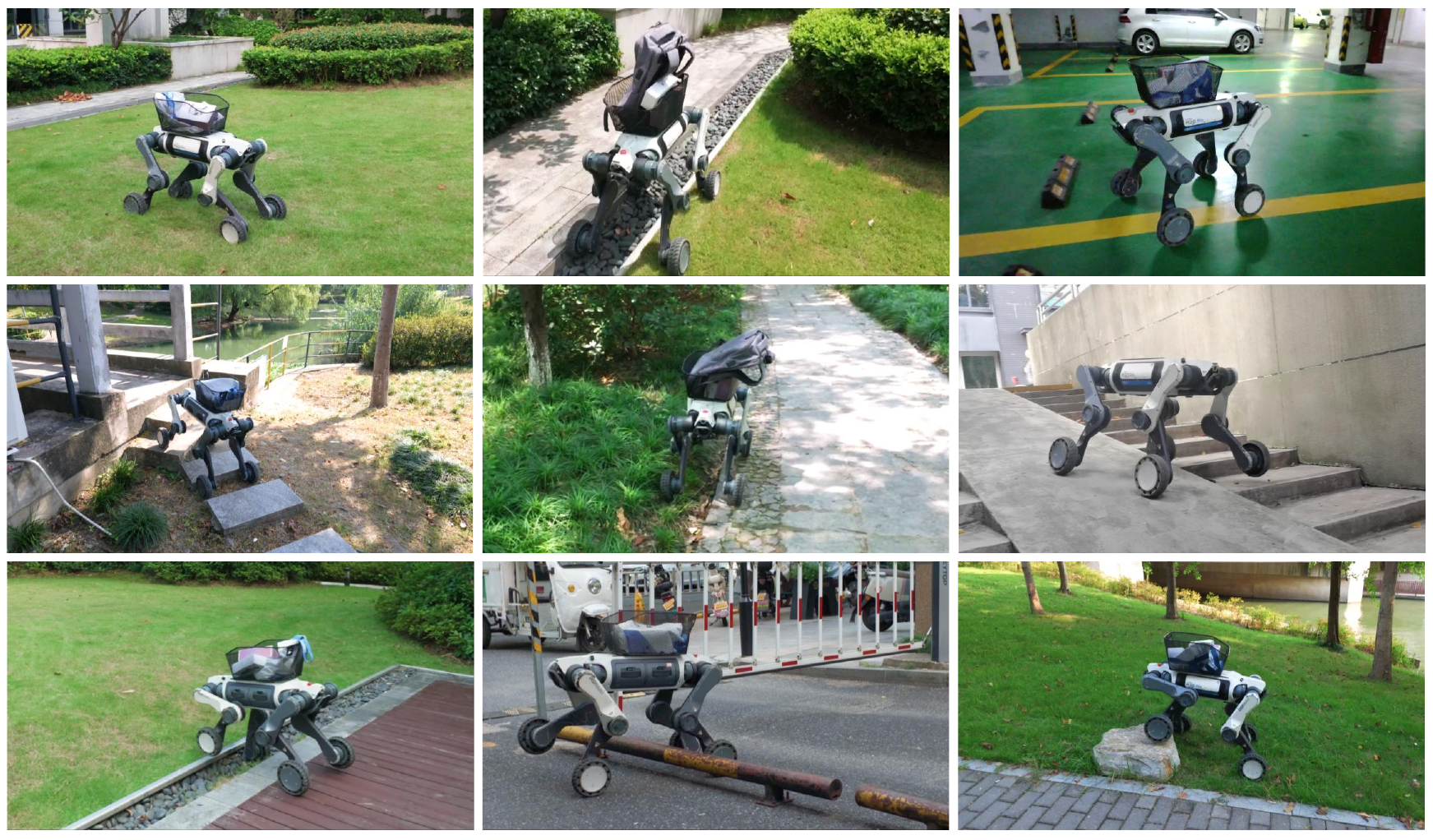}
\vspace{-1.6em}
\caption{Real-world deployment of LocoSusp on diverse outdoor and semi-structured environments, including grass, stone roads, stepping stones, speed bumps, and terrain transitions.}
\label{fig:real_robot_demo}
\vspace{-0.6em}
\end{figure}

We further validate LocoSusp on the M20 wheeled-legged platform under diverse real-world terrain disturbances (Fig. ~\ref{fig:real_robot_demo}). As shown in Figure~\ref{fig:real_robot_with_data}, the robot is evaluated on representative scenarios involving discrete impact and terrain transition, corresponding to different excitation characteristics considered in simulation. We evaluate stability from three views: onboard RGB camera stability, torso IMU/kinematic logs, and payload motion. The qualitative results show that LocoSusp produces smaller camera shake, lower vertical acceleration, and smoother payload motion while preserving traversability. These demonstrate that the learned suspension behavior transfers to physical environments, maintaining stable body motion and onboard camera stability under real contact uncertainties.

\section{Conclusion}

In this paper, we proposed LocoSusp, a terrain-adaptive active suspension framework for stable wheeled-legged mobile bases. The key idea is to reinterpret the wheel-leg body as a mobile suspension platform: terrain traversal remains necessary, but the robot must also suppress vibration, reduce impact, and maintain usable payload and camera stability. LocoSusp decomposes control into a stability backbone and terrain-conditioned residual suspension adaptation, uses contact-driven terrain excitation features to enable adaptive whole-body responses under diverse terrain disturbances. Simulation and M20 real-robot demonstrations validate the framework across diverse terrain conditions, demonstrating its capability for reliable mobile platform operation.

\bibliographystyle{IEEEtran}
\bibliography{references}

\end{document}